\documentclass[11pt,a4paper]{article}
\usepackage[hyperref]{acl2021}
\usepackage{times}
\usepackage{latexsym}

\usepackage{microtype}

\usepackage[aboveskip=5pt]{caption}
\usepackage{subcaption}

\usepackage{enumitem}
\newlist{compactitem}{itemize}{3} 
\setlist[compactitem]{label=\textbullet, nosep}

\usepackage{amssymb}
\usepackage{pifont}
\newcommand{\cmark}{\ding{51}}%
\newcommand{\xmark}{\ding{55}}%

\usepackage{pbox}
\usepackage{hyperref}
\usepackage{url}
\usepackage{comment}
\usepackage[flushleft]{threeparttable}
\usepackage{graphicx}
\usepackage{amsmath}
\usepackage{tabularx}
\usepackage{multirow, makecell}
\usepackage{algorithm}
\usepackage{xcolor}
\usepackage{booktabs}
\usepackage{paralist}
\DeclareMathOperator*{\softmax}{softmax}

\usepackage{cleveref}
\crefname{section}{§}{§§}
\Crefname{section}{§}{§§}

\aclfinalcopy 
\def\aclpaperid{3236} 

\definecolor{MyColor}{RGB}{50, 100, 250}
\definecolor{Orange}{RGB}{244, 101, 66}
\definecolor{Red}{RGB}{255, 0, 0}
\definecolor{Green}{RGB}{0, 255, 0}
\definecolor{Blue}{RGB}{0, 0, 255}

\newcommand{\mytilde}{\raise.17ex\hbox{$\scriptstyle\mathtt{\sim}$}}

\usepackage{xcolor,colortbl}

\definecolor{gray}{RGB}{171, 178, 185}
\definecolor{lgray}{RGB}{204, 209, 209}

\usepackage{arydshln}
\title{Syntax-augmented Multilingual BERT for Cross-lingual Transfer}

\author{
Wasi Uddin Ahmad$^\dagger$\thanks{~~Work done during internship at Facebook AI.}, Haoran Li$^\ddagger$, Kai-Wei Chang$^\dagger$,  Yashar Mehdad$^\ddagger$ \\
$^\dagger$University of California, Los Angeles, $^\ddagger$Facebook AI \\
$^\dagger$\texttt{\{wasiahmad,kwchang\}@cs.ucla.edu},  $^\ddagger$\texttt{\{aimeeli,mehdad\}@fb.com}
}

\date{}

\begin{document}

\setlength{\abovedisplayskip}{5pt}
\setlength{\belowdisplayskip}{5pt}

\maketitle

\begin{abstract}
In recent years, we have seen a colossal effort in pre-training multilingual text encoders using large-scale corpora in many languages to facilitate cross-lingual transfer learning. However, due to typological differences across languages, the cross-lingual transfer is challenging. Nevertheless, language syntax, e.g., syntactic dependencies, can bridge the typological gap. Previous works have shown that pre-trained multilingual encoders, such as mBERT \cite{devlin-etal-2019-bert}, capture language syntax, helping cross-lingual transfer. This work shows that explicitly providing language syntax and training mBERT using an auxiliary objective to encode the universal dependency tree structure helps cross-lingual transfer. We perform rigorous experiments on four NLP tasks, including text classification, question answering, named entity recognition, and task-oriented semantic parsing. The experiment results show that syntax-augmented mBERT improves cross-lingual transfer on popular benchmarks, such as PAWS-X and MLQA, by 1.4 and 1.6 points on average across all languages. In the \emph{generalized} transfer setting, the performance boosted significantly, with 3.9 and 3.1 points on average in PAWS-X and MLQA.

\end{abstract}

\section{Introduction}

\begin{figure}[!ht]
\centering
\hspace*{-2mm}
\includegraphics[width=1.05\linewidth]{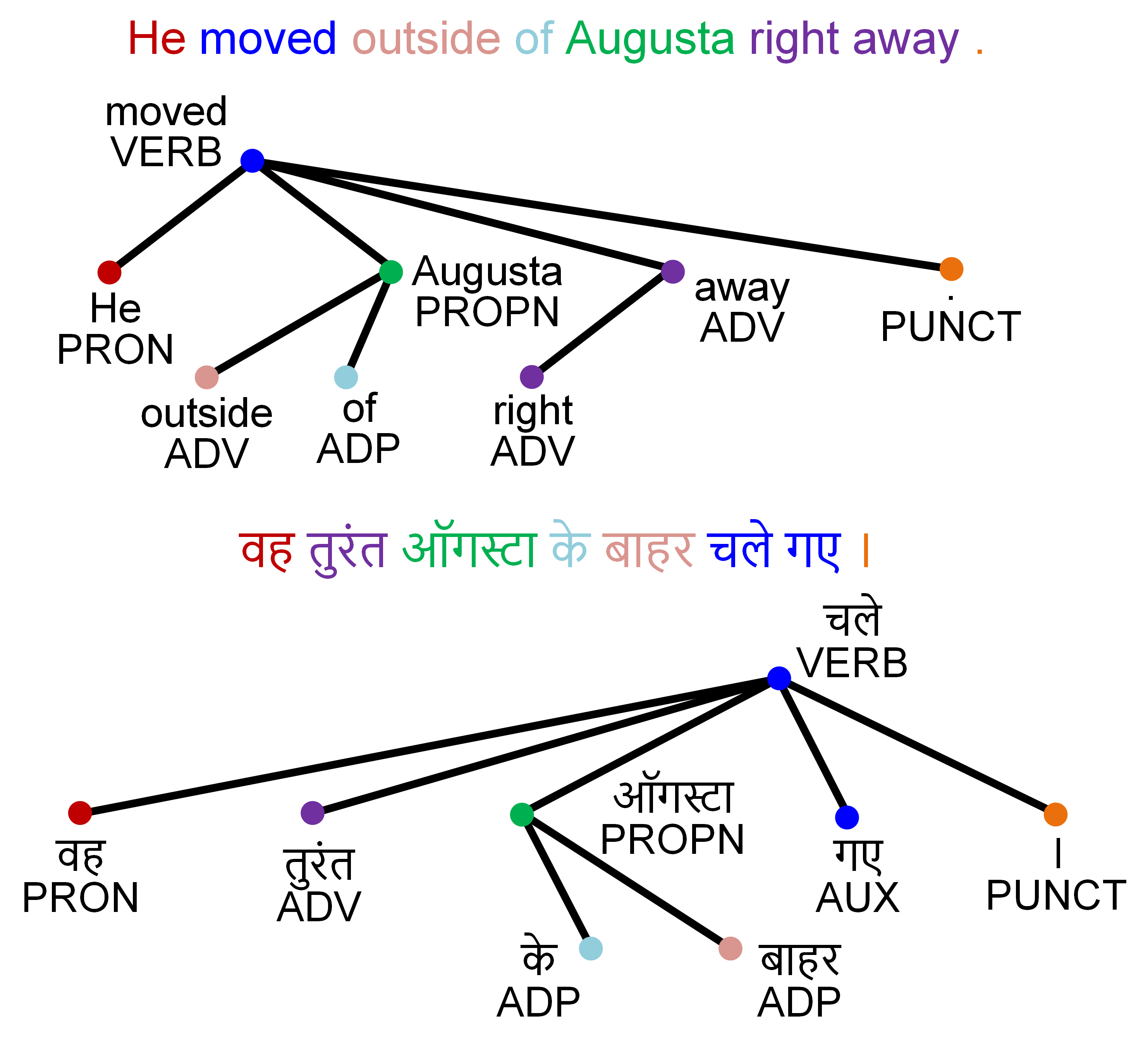}
\caption{
Two parallel sentences in English and Hindi from XNLI \cite{conneau-etal-2018-xnli} dataset.
The words highlighted with the same color have the same meaning.
Although the sentences have a different word order, their syntactic dependency structure is similar.
}
\label{fig:example}
\end{figure}

Cross-lingual transfer reduces the requirement of labeled data to perform natural language processing (NLP) in a target language, and thus has the ability to avail NLP applications in low-resource languages.
However, transferring across languages is challenging because of linguistic differences at levels of morphology, syntax, and semantics.
For example, word order difference is one of the crucial factors that impact cross-lingual transfer \cite{ahmad-etal-2019-difficulties}.
The two sentences in English and Hindi, as shown in Figure \ref{fig:example} have the same meaning but a different word order (while English has an SVO ({\tt Subject-Verb-Object}) order, Hindi follows SOV).
However, the sentences have a similar dependency structure, and the constituent words have similar part-of-speech tags.
Presumably, language syntax can help to bridge the typological differences across languages.

\begin{figure*}[t]
\centering
\resizebox{\linewidth}{!}{%
\def\arraystretch{1.1}%
\begin{tabular}{l | l | p{0.75\linewidth}}
\hline
\multirow{2}{*}{Q} & English & How many {\color{red}members} of the {\color{brown}Senate} are {\color{blue}elected}? \\ 
\cdashline{2-3}
& Spanish & Cuántos {\color{red}miembros} del {\color{brown}Senado} son {\color{blue}elegidos}?  \\ \hline
\multirow{4}{*}{C} & \multirow{2}{*}{English} & The Chamber of Deputies has 630 {\color{blue}elected} {\color{red}members}, while the {\color{brown}Senate} has 315 {\color{blue}elected} {\color{red}members}. \ldots \\ 
\cdashline{2-3}
& \multirow{2}{*}{Spanish} & La cámara de los diputados está formada por 630 {\color{red}miembros}, mientras que hay 315 {\color{purple}senadores} más los {\color{purple}senadores} vitalicios. \ldots \\ \hline
\multirow{4}{*}{A} & \multirow{2}{*}{mBERT} & [Q:English-C:English] 315 (\cmark); [Q:Spanish-C:Spanish] 630 (\xmark) \\
&  &  [Q:Spanish-C:English] 315 (\cmark); [Q:English-C:Spanish] 630 (\xmark) \\ 
\cdashline{2-3}
& \multirow{2}{*}{mBERT + Syn.} &  [Q:English-C:English] 315 (\cmark); [Q:Spanish-C:Spanish] 315 (\cmark) \\
&  &  [Q:Spanish-C:English] 315 (\cmark); [Q:English-C:Spanish] 315 (\cmark) \\
\hline
\end{tabular}
}
\caption{
A parallel QA example in English (en) and Spanish (es) from MLQA \cite{lewis-etal-2020-mlqa} with predictions from mBERT and our proposed syntax-augmented mBERT.
In ``Q:x-C:y'', x and y indicates question and context languages, respectively.
Based on our analysis of the highlighted tokens' attention weights, we conjecture that mBERT answers 630 as the token is followed by ``miembros'', while 315 is followed by ``senadores'' in Spanish.
}
\label{table:qa_ex1}
\end{figure*}

In recent years, we have seen a colossal effort to pre-train Transformer encoder \cite{vaswani2017attention} on large-scale unlabeled text data in one or many languages.
Multilingual encoders, such as mBERT \cite{devlin-etal-2019-bert} or XLM-R \cite{conneau-etal-2020-unsupervised} map text sequences into a shared multilingual space by jointly pre-training in many languages.
This allows us to transfer the multilingual encoders across languages
and have found effective for many NLP applications, 
including text classification \cite{bowman-etal-2015-large, conneau-etal-2018-xnli}, question answering \cite{rajpurkar-etal-2016-squad, lewis-etal-2020-mlqa}, named entity recognition \cite{pires-etal-2019-multilingual, wu-dredze-2019-beto}, and more.
Since the introduction of mBERT, several works \cite{wu-dredze-2019-beto, pires-etal-2019-multilingual, K2020Cross-Lingual} attempted to reason their success in cross-lingual transfer.
In particular, \citet{wu-dredze-2019-beto} showed that mBERT captures language syntax that makes it effective for cross-lingual transfer.
A few recent works \cite{hewitt-manning-2019-structural, jawahar-etal-2019-bert, chi-etal-2020-finding} suggest that BERT learns compositional features; mimicking a tree-like structure that agrees with the Universal Dependencies taxonomy.

However, fine-tuning for the downstream task in a source language may not require mBERT to retain structural features or learn to encode syntax. 
We argue that encouraging mBERT to learn the correlation between syntax structure and target labels can benefit cross-lingual transfer.
To support our argument, we show an example of question answering (QA) in Figure \ref{table:qa_ex1}.
In the example, mBERT predicts incorrect answers given the Spanish language context that can be corrected by exploiting syntactic clues.
Utilizing syntax structure can also benefit \emph{generalized} cross-lingual transfer \cite{lewis-etal-2020-mlqa} where the input text sequences belong to different languages. 
For example, answering an English question based on a Spanish passage or predicting text similarity given the two sentences as shown in Figure \ref{fig:example}.
In such a setting, syntactic clues may help to align sentences.

In this work, we propose to augment mBERT with universal language syntax while fine-tuning on downstream tasks.
We use a graph attention network (GAT) \cite{velickovic2018graph} to learn structured representations of the input sequences that are incorporated into the self-attention mechanism.
We adopt an auxiliary objective to train GAT such that it embeds the dependency structure of the input sequence accurately.
We perform an evaluation on \emph{zero-shot} cross-lingual transfer for text classification, question answering, named entity recognition, and task-oriented semantic parsing.
Experiment results show that augmenting mBERT with syntax improves cross-lingual transfer, such as in PAWS-X and MLQA, by 1.4 and 1.6 points on average across all the target languages. 
Syntax-augmented mBERT achieves remarkable gain in the generalized cross-lingual transfer; in PAWS-X and MLQA, performance is boosted by 3.9 and 3.1 points on average across all language pairs. 
Furthermore, we discuss challenges and limitations in modeling universal language syntax.
We release the code to help future works.\footnote{https://github.com/wasiahmad/Syntax-MBERT}

\section{Syntax-augmented Multilingual BERT}

Multilingual BERT (mBERT) \cite{devlin-etal-2019-bert} enables cross-lingual learning as it embeds text sequences into a shared multilingual space.
mBERT is fine-tuned on downstream tasks, e.g., text classification using \emph{monolingual} data and then directly employed to perform on the target languages.
This refers to \emph{zero-shot} cross-lingual transfer.
Our main idea is to augment mBERT with language syntax for zero-shot cross-lingual transfer.
We employ graph attention network (GAT) \cite{velickovic2018graph} to learn syntax representations and fuse them into the self-attention mechanism of mBERT.

In this section, we first briefly review the transformer encoder that bases mBERT (\cref{sec:trans_enc}), and then describe the graph attention network (GAT) that learns syntax representations from dependency structure of text sequences (\cref{sec:gat}).
Finally, we describe how language syntax is explicitly incorporated into the transformer encoder (\cref{sec:syntax-bert}).

\subsection{Transformer Encoder}
\label{sec:trans_enc}


Transformer encoder \cite{vaswani2017attention} is composed of an embedding layer and stacked encoder layers.
Each encoder layer consists of two sublayers, a multi-head attention layer followed by a fully
connected feed-forward layer.
We detail the process of encoding an input token sequence ($w_1, \ldots, w_n$) into a sequence of vector representations $H = [h_1, \ldots, h_n]$ as follows.

\paragraph{Embedding Layer}
is parameterized by two embedding matrices --- the token embedding matrix  $W_e \in R^{U \times d_{model}}$ and the position embedding matrix $W_p \in R^{U \times d_{model}}$ (where $U$ is the vocabulary size and $d_{model}$ is the encoder output dimension). An input text sequence enters into the model as two sequences: the token sequence ($w_1, \ldots, w_n$) and the corresponding absolute position sequence  ($p_1, \ldots, p_n$). 
The output of the embedding layer is a sequence of vectors $\{x_i\}_{i=1}^n$ 
where $x_i = w_i W_e + p_i W_p$.
The vectors are packed into matrix $H^0=[x_1, \ldots, x_{n}] \in R^{n \times d_{model}}$ and fed to an $L$-layer encoder.

\paragraph{Multi-head Attention} 
allows to jointly attend to information from different representation subspaces, known as \emph{attention heads}. 
Multi-head attention layer composed of $h$  
attention heads with the same parameterization structure.
At each attention head, the output from the previous layer $H^{l-1}$ is first linearly projected into  queries, keys, and values as follows.
\begin{gather*}
Q = H^{l-1}W^Q_l, K = H^{l-1}W^K_l, V = H^{l-1}W^V_l,
\end{gather*}
where the parameters $W^Q_l, W^K_l \in R^{d_{model} \times d_k}$ and $W^V_l \in R^{d_{model} \times d_v}$ are unique per attention head.
Then scaled dot-product attention is performed to compute the output vectors $\{o_i\}_{i=1}^{n} \in R^{n \times d_v}$.
\begin{equation}
\label{eq:attention}
\begin{split}
    & \textrm{Attention}(Q, K, V, M, d_k) \\
    & = \softmax\left(\frac{QK^T + M}{\sqrt{d_k}}\right) V,
\end{split}
\end{equation}
where  $M \in \mathbb{R}^{n\times n}$ is the masking matrix that determines whether a pair of input positions can attend each other. 
In classic multi-head attention, $M$ is a \emph{zero} matrix (all positions can attend each other).

The output vectors from all the attention heads are concatenated and projected into $d_{model}$ dimension using the parameter matrix $W_o \in R^{hd_v \times d_{model}}$.
Finally the vectors are passed through a feed-forward network to output $H^l \in R^{n \times d_{model}}$.


\subsection{Graph Attention Network}
\label{sec:gat}
\begin{figure}
\centering
\includegraphics[width=1.0\linewidth]{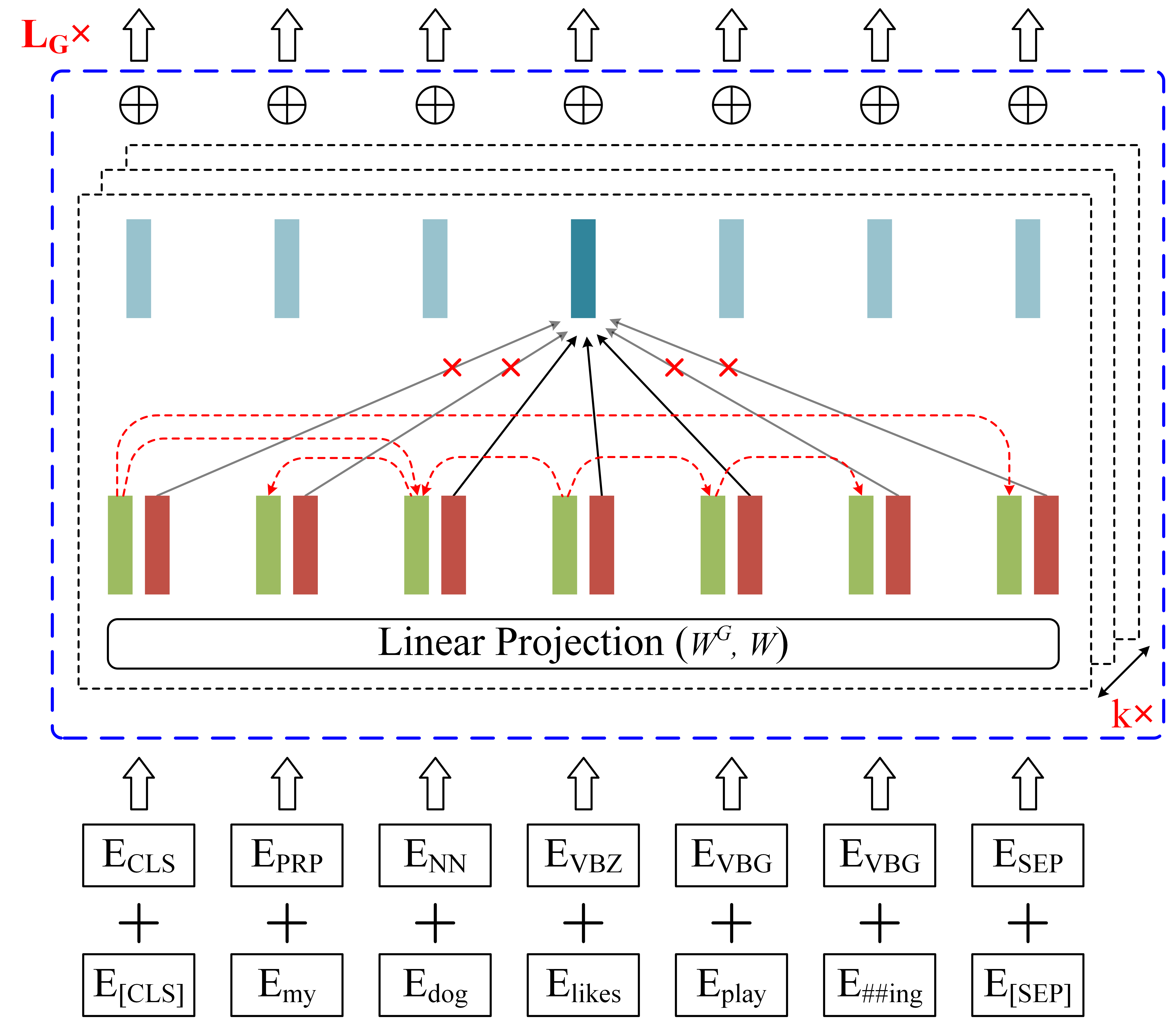}
\vspace{1mm}
\vspace{-3mm}
\caption{
A simplified illustration of the multi-head self-attention in the graph attention network wherein each head attention is allowed between words within $\delta$ distance from each other in the dependency graph.
For example, as shown, in one of the attention heads, the word ``likes'' is only allowed to attend its adjacent ($\delta$=1) words ``dog'' and ``play''.
}
\label{figure:gat}
\end{figure}

We embed the syntax structure of the input token sequences using their universal dependency parse.
A dependency parse is a directed graph where the nodes represent words, and the edges represent dependencies (the dependency relation between the head and dependent words).
We use a graph attention network (GAT) \cite{velickovic2018graph} to embed the dependency tree structure of the input sequence. 
We illustrate GAT in Figure \ref{figure:gat}.

Given the input sequence, the words ($w_i$) and their part-of-speech tags ($pos_i$) are embedded into vectors using two parameter matrices: the token embedding matrix $W_e$ and the part-of-tag embedding $W_{pos}$. The input sequence is then encoded into an input matrix $\mathcal{G}^0 = [g_1, \ldots, g_n]$, where $g_i = w_i W_e + pos_i W_{pos} \in R^{d_{model}}$.
Note that token embedding matrix $W_e$ is shared between GAT and the Transformer encoder.
Then $\mathcal{G}^0$ is fed into an $L_\mathcal{G}$-layer GAT where each layer generates word representations by attending their adjacent words. 
GAT uses the multi-head attention mechanism and perform a \emph{dependency-aware} self-attention as
\begin{gather}
\label{eq:dependency_attention}
    \mathcal{O} = \textrm{Attention}(\mathcal{T}, \mathcal{T}, V, M, d_g)
\end{gather}
namely setting the query and key matrices to be the same $\mathcal{T} \in R^{n \times d_g}$ respectively 
and the mask $M$ by 
\begin{gather}
\label{eq:mask}
M_{ij} = \begin{cases}
        0, & D_{ij} \leq \delta \\
        -\infty, & \text{otherwise}
    \end{cases}
\end{gather}
where $D$ is the distance matrix and $D_{ij}$ indicates the shortest path distance between word $i$ and $j$ in the dependency graph structure.

Typically in GAT, $\delta$ is set to 1; allowing attention between adjacent words only.
However, in our study, we find setting $\delta$ to [2, 4] helpful for the downstream tasks.
Finally, the vector representations from all the attention heads (as in Eq. \eqref{eq:dependency_attention}) are concatenated to form the output representations $\mathcal{G}^l \in R^{n \times kd_g}$, where $k$ is the number of attention heads employed.
The goal of the GAT encoder is to encode the dependency structure into vector representations. 
Therefore, we design GAT to be light-weight; consisting of much less parameters in comparison to Transformer encoder.
Note that, GAT does not employ positional representations and only consists of multi-head attention; there is no feed-forward sublayer and residual connections.

\paragraph{Dependency Tree over Wordpieces and Special Symbols}
mBERT tokenizes the input sequence into subword units, also known as wordpieces.
Therefore, we modify the dependency structure of linguistic tokens to accommodate wordpieces.
We introduce additional dependencies between the first subword (head) and the rest of the subwords (dependents) of a linguistic token.
More specifically, we introduce new edges from the head subword to the dependent subwords.
The inputs to mBERT use special symbols: [CLS] and [SEP].
We add an edge from the [CLS] token to the root of the dependency tree and the [SEP] tokens.

\subsection{Syntax-augmented Transformer Encoder}
\label{sec:syntax-bert}
We want the Transformer encoder to consider syntax structure while performing the self-attention between input sequence elements.
We use the syntax representations produced by GAT (outputs from the last layer, denoting as $\mathcal{G}$) to \emph{bias} the self-attention.
\begin{gather*}
    O = \textrm{Attention}(Q + \mathcal{G} G^Q_l, K + \mathcal{G} G^K_l, V, M, d_k),
\end{gather*}
where $G^Q_l, G^K_l \in R^{d_{kd_g} \times d_k}$ are new parameters that learn representations to bias the self-attention.
We consider the addition terms ($\mathcal{G} G^Q_l, \mathcal{G} G^K_l$) as syntax-bias that provide syntactic clues to guide the self-attention.
The high-level intuition behind the syntax bias is to attend tokens with a specific part-of-speech tag sequence or dependencies.\footnote{In example shown in Figure \ref{table:qa_ex1}, token dependencies: [en: root $\rightarrow$ has $\rightarrow$ has $\rightarrow$ members $\rightarrow$ 315], and [es: root $\rightarrow$ formada $\rightarrow$ hay $\rightarrow$ senadores $\rightarrow$ 315]  or corresponding part-of-speech tag sequence [VERB $\rightarrow$ VERB $\rightarrow$ NOUN $\rightarrow$ NUM]) may help mBERT to predict the correct answer.}


\paragraph{Syntax-heads}
mBERT employs $h$ (=12) attention heads and the syntax representations can be infused into one or more of these heads, and we refer them as \emph{syntax-heads}.
In our experiments, we observed that instilling structural information into many attention heads degenerates the performance.
For the downstream tasks, we consider one or two syntax-heads that gives the best performance.\footnote{This aligns with the findings of \citet{hewitt-manning-2019-structural} as they showed 64 or 128 dimension of the contextual representations are sufficient to capture the syntax structure.}

\paragraph{Syntax-layers} 
refers to the encoder layers that are infused by syntax representations from GAT. 
mBERT has a 12-layer encoder and our study finds considering all of the layers as \emph{syntax-layers} beneficial for cross-lingual transfer.


\begin{table*}[t]
\centering
\begin{tabular}{l c r r r c c}
\hline
Dataset & Task & $|$Train$|$ & $|$Dev$|$ & $|$Test$|$  & $|$Lang$|$ & Metric \\
\hline
XNLI & Classification & 392K & 2.5K & 5K & 13 & Accuracy \\
PAWS-X & Classification & 49K & 2K & 2K & 7 & Accuracy \\ 
\hline
MLQA & QA & 87K & 34K & 4.5K-11K & 7 & F1 / Exact Match \\
XQuAD & QA & 87K & 34K & 1190 & 10 & F1 / Exact Match \\
\hline
Wikiann & NER & 20K & 10K & 1K-10K & 15 & F1\\ 
CoNLL & NER & 15K & 2K-3K & 1.5K-5K & 4 & F1\\ 
\hline
mTOP & Semantic Parsing & 15.7K & 2.2K & 2.8K-4.4K & 5 & Exact Match \\
mATIS++ & Semantic Parsing & 4.5K & 490 & 893 & 9 & Exact Match \\
\hline
\end{tabular}
\caption{Statistics of the evaluation datasets. $|$Train$|$, $|$Dev$|$ and $|$Test$|$ are the numbers of examples in the training, dev and test sets, respectively. 
For train set, the number is for the source language, English, while for dev and test set, the number is for each target language. $|$Lang$|$ is the number of target languages we consider for each task. 
}
\label{table:statitics}
\end{table*}

\subsection{Fine-tuning}
\label{sec:fine-tuning}
We jointly fine-tune mBERT and GAT on downstream tasks in the source language (English in this work) following the standard procedure.
However, the task-specific training may not guide GAT to encode the tree structure.
Therefore, we adopt an auxiliary objective that supervises GAT to learn representations which can be used to decode the tree structure.
More specifically, we use GAT's output representations $\mathcal{G} = [g_1, \ldots, g_n]$ to predict the tree distance between all pairs of words $(g_i, g_j)$ and the tree depth $||g_i||$ of each word $w_i$ in the input sequence.
Following \citet{hewitt-manning-2019-structural}, we apply a linear transformation $\theta_1 \in R^{m \times kd_g}$ to compute squared distances as follows. 
\begin{gather*}
    d_{\theta_1} (g_i, g_j)^2 = (\theta_1(g_i - g_j))^T (\theta_1(g_i - g_j)).
\end{gather*}
The parameter matrix $\theta_1$ is learnt by minimizing:
\begin{gather*}
    \min\limits_{\theta_1} \sum_s \frac{1}{n^2} \sum_{i, j} |dist(w_i, w_j)^2 - d_{\theta} (g_i, g_j)^2 |,
\end{gather*}
where $s$ denotes all the text sequences in the training corpus.
Similarly, we train another parameter matrix $\theta_2$ 
to compute squared vector norms,  $d_{\theta_2}(g_i) = (\theta_2 g_i)^T(\theta_2 g_i)$ that characterize the tree depth of the words.
We train GAT's parameters and $\theta_1, \theta_2$ by minimizing the  loss: $\mathcal{L} = \mathcal{L}_{task} + \alpha (\mathcal{L}_{dist} +  \mathcal{L}_{depth})$, where $\alpha$ is weight for the tree structure prediction loss.

\begin{table*}[ht!]
	\centering
	\resizebox{\linewidth}{!}{%
	\small
	\def\arraystretch{1.2}%
	\setlength{\tabcolsep}{2.5pt}
	\begin{tabular}{@{}l|c c c c c c c c c c c c c c c c c|r@{}}
	\hline
	Model	&	en	&	ar	&	bg	&	de	&	el	&	es	&	fr	&	hi	&	ru	&	tr	&	ur	&	vi	&	zh	&	ko	&	ja	&	nl	&	pt  &   AVG	\\ \hline
	\multicolumn{19}{@{}l}{Classification - XNLI \cite{conneau-etal-2018-xnli}}
	\\ \hline
    [1]	&	80.8	&	64.3	&	68.0	&	70.0	&	65.3	&	73.5	&	73.4	&	58.9	&	67.8	&	60.9	&	57.2	&	69.3	&	67.8	&	-	&	-	&	-	&	-	&   67.5	\\
	mBERT	&	81.8	&	63.8	&	68.0	&	70.7	&	65.4	&	73.8	&	72.4	&	59.3	&	68.4	&	60.7	&	56.7	&	68.6	&	67.8	&	-	&	-	&	-	&	-	&   67.5	\\
	\, + Syn.	&	81.6	&	{\bf 65.4}	&	{\bf 69.3}	&	70.7	&	{\bf 66.5}	&	{\bf 74.1}	&	{\bf 73.2}	&	{\bf 60.5}	&	68.8	&	{\bf 62.4}	&	{\bf 58.7}	&	{\bf 69.9}	&	{\bf 69.3}	&	-	&	-	&	-	&	-	&   {68.5}	\\ \hline
	\multicolumn{19}{@{}l}{Classification - PAWS-X \cite{yang-etal-2019-paws}}
	\\ \hline
	[1]	&	94.0	&	-	&	-	&	85.7	&	-	&   87.4	&	87.0	&	-	&	-	&	-	&	-	& -	&	77.0	&	69.6	&	73.0	&	-	&	-	&   82.0	\\
	mBERT	&	93.9	&	-	&	-	&	85.7	&	-	&   88.4	&	87.6	&	-	&	-	&	-	&	-	&	-   &   78.0	&	73.6	&	73.1	&	-	&	-	&   82.9	\\
	\, + Syn.	&	94.0	&	-	&	-	&	85.9    &   -	&	{\bf 89.1}	&	{\bf 88.2}	&	-	&	-	&	-	&	-	&	-   &   {\bf 80.7}	&	{\bf 76.3}	&	{\bf 75.8}	&	-	&	-	&   {84.3}	\\ \hline
	\multicolumn{18}{@{}l}{NER - Wikiann \cite{pan-etal-2017-cross}}
	\\ \hline
    [1]	&	85.2	&	41.1	&	77.0	&	78.0	&	72.5	&	77.4	&	79.6	&	65.0	&	64.0	&	71.8	&	36.9	&	71.8	&	-	&	59.6	&	-	&	81.8	&	80.8	&   {69.5}	\\
	mBERT	&	83.6	&	38.8	&	77.0	&	76.0	&	70.4	&	74.7	&	78.9	&	63.4	&	63.5	&	70.9	&	37.7	&	73.5	&	-	&	59.3	&	-	&	81.9	&	78.7	&   68.5	\\
	\, + Syn.	&	84.4	&	{\bf 40.0}	&	77.0	&	{\bf 77.0}	&	{\bf 71.5}	&	{\bf 76.1}	&	79.3	&	64.2	&	63.8	&	71.4	&	37.3	&	72.7	&	-	&	59.3	&	-	&	81.9	&	79.0	&   {69.0}	\\ \hline
	\multicolumn{19}{@{}l}{NER - CoNLL \cite{tjong-kim-sang-2002-introduction,tjong-kim-sang-de-meulder-2003-introduction}}
	\\ \hline
    [2]	&	90.6	&	-	&	-	&	69.2	&	-	&	75.4	&	-	&	-	&	-	&	-	&	-	&	-	&	-	&	-	&	-	&	77.9	&	-	&   {78.2}	\\
	mBERT	&	90.7	&	-	&	-	&	68.3	&	-	&	74.5	&	-	&	-	&	-	&	-	&	-	&	-	&	-	&	-	&	-	&	77.6	&	-	&   77.8	\\
	\ + Syn.	&	90.6	&	-	&	-	&	{\bf 69.1}	&	-	&	73.6	&	-	&	-	&	-	&	-	&	-	&	-	&	-	&	-	&	-	&	{\bf 78.5}	&	-	&   {78.0}	\\ \hline
	\multicolumn{19}{@{}l}{QA - MLQA \cite{lewis-etal-2020-mlqa}}
	\\ \hline
    [3]	&	77.7	&	45.7	&	-	&	57.9	&	-	&	64.3	&	-	&	43.8	&	-	&	-	&	-	&	57.1	&	57.5	&	-	&	-	&	-	&	-	&   57.7	\\
	mBERT	&	80.5	&	47.2	&	-	&	59.0	&	-	&	63.9	&	-	&	47.5	&	-	&	-	&	-	&	56.5	&	56.6	&	-	&	-	&	-	&	-	&   58.7	\\
	\, + Syn.	&	80.4	&	{\bf 48.9}	&	-	&	{\bf 60.8}	&	-	&	{\bf 65.9}	&	-	&	46.7	&	-	&	-	&	-	&	{\bf 59.3}	&	{\bf 60.1}	&	-	&	-	&	-	&	-	&   {60.3}	\\ \hline
	\multicolumn{19}{@{}l}{QA - XQuAD \cite{artetxe-etal-2020-cross}}
	\\ \hline
    [1]	&	83.5	&	61.5	&	-	&	70.6	&	62.6	&	75.5	&	-	&	59.2	&	71.3	&	55.4	&	-	&	69.5	&	58.0	&	-	&	-	&	-	&	-	&   {66.7}	\\
	mBERT	&	84.2	&	54.8	&	-	&	68.9	&	60.2	&	71.1	&	-	&	55.7	&	68.6	&	48.9	&	-	&	64.0	&	57.2	&	-	&	-	&	-	&	-	&   63.4	\\
	\, + Syn.	&	84.0	&	55.5	&	-	&	{\bf 71.4}	&	{\bf 61.3}	&	{\bf 72.8}	&	-	&	54.6	&	68.4	&	{\bf 49.8}	&	-	&	{\bf 67.6}	&	56.1	&	-	&	-	&	-	&	-	&   {64.2}	\\ \hline
	\multicolumn{19}{@{}l}{Semantic Parsing - mTOP \cite{li2020mtop}}
	\\ \hline
    mBERT	&	81.0	&	-	&	-	&	28.1	&	-	&	40.2	&	38.8    &	9.8	&	-	&	-	&	-	&	-	&	-	&	-	&	-	&	-	&	-	&   39.6	\\
	\, + Syn.	&	81.3	&	-	&	-	&	{\bf 30.0}	&	-	&	{\bf 43.0}	&	{\bf 41.2}	&	{\bf 11.5}	&	-	&	-	&	-	&	-	&	-	&	-	&	-	&	-	&	-	&   {41.4}	\\ \hline
	\multicolumn{19}{@{}l}{Semantic Parsing - mATIS++ \cite{xu-etal-2020-end}}
	\\ \hline
	mBERT   &	86.0	&	-	&	-	&	38.1	&	-	&	43.7	&	36.9	&	16.2	&	-	&	1.3	&	-	&	-	&	7.8	&	-	&	28.2	&	-  &	38.2	&   32.9	\\
	\, + Syn.	&	86.2	&	-	&	-	&   {\bf 40.1}	&	-	&	{\bf 44.5}	&	{\bf 38.9}	&	{\bf 18.7}	&	-	&	1.5	&	-	&	-	&	8.0	&	-	&	27.3	&	-	&	37.3	&   {33.6}	\\
		\hline
	\end{tabular}
}
\caption{
Cross-lingual transfer results for all the evaluation tasks (on test set) across 17 languages.
We report F1 score for the question answering (QA) datasets (for other datasets, see Table \ref{table:statitics}). 
We train and evaluate mBERT on the same pre-processed datasets and considers its performance as the \emph{baseline} (denoted by ``mBERT'' rows in the table) for syntax-augmented mBERT (denoted by ``+ Syn.'' rows in the table).
Bold-faced values indicate that the syntax-augmented mBERT is statistically significantly better (by paired bootstrap test, p $<$ 0.05) than the \emph{baseline}.
We include results from published works ([1]: \citet{hu2020xtreme}, [2]: \citet{liang2020xglue}, and [3]: \citet{lewis-etal-2020-mlqa}) as a reference.
Except for the QA datasets, all our results are averaged over three different seeds.
}
\label{table:all_result} 
\end{table*}

\paragraph{Pre-training GAT}
Unlike mBERT's parameters, GAT's parameters are trained from scratch during task-specific fine-tuning.
For low-resource tasks, GAT may not learn to encode the syntax structure accurately. 
Therefore, we utilize the universal dependency parses\footnote{https://universaldependencies.org/} to pre-train GAT on the source and target languages.
Note that, the pre-training objective for GAT is to predict the tree distances and depths as described above.

\section{Experiment Setup}
To study syntax-augmented mBERT's performance in a broader context, we perform an evaluation on four NLP applications: text classification, named entity recognition, question answering, and task-oriented semantic parsing. 
Our evaluation focuses on assessing the usefulness of utilizing universal syntax in the \emph{zero-shot} cross-lingual transfer.

\subsection{Evaluation Tasks}

\paragraph{Text Classification}
We conduct experiments on two widely used cross-lingual text classification tasks: (i) natural language inference and (ii) paraphrase detection. 
We use the XNLI \cite{conneau-etal-2018-xnli} and PAWS-X \cite{yang-etal-2019-paws} datasets for the tasks, respectively.
In both tasks, a pair of sentences is given as input to mBERT. 
We combine the dependency tree structure of the two sentences by adding two edges from the [CLS] token to the roots of the dependency trees.
A linear classifier takes the contextual representation for the [CLS] token to predict the target label.

\paragraph{Named Entity Recognition}
is a structure prediction task that requires to identify the named entities mentioned in the input sentence.
We use the Wikiann dataset \cite{pan-etal-2017-cross} and a subset of two tasks from CoNLL-2002 \cite{tjong-kim-sang-2002-introduction} and CoNLL-2003 NER \cite{tjong-kim-sang-de-meulder-2003-introduction}.
We collect the CoNLL datasets from XGLUE \cite{liang2020xglue}.
In both datasets, there are 4 types of named entities: Person, Location, Organization, and Miscellaneous.\footnote{Miscellaneous entity type covers named entities that do not belong to the other three types}

\paragraph{Question Answering}
We evaluate on two cross-lingual question answering benchmarks, MLQA \cite{lewis-etal-2020-mlqa}, and XQuAD \cite{artetxe-etal-2020-cross}. 
We use the SQuAD dataset \cite{rajpurkar-etal-2016-squad} for training and validation.
In the QA task, the inputs are a question and a context passage that consists of many sentences.
We formulate QA as a multi-sentence reading comprehension task; jointly train the models to predict the answer sentence and extract the answer span from it.
We concatenate the question and each sentence from the context passage and use the [CLS] token representation to score the candidate sentences.
We provide more details of the QA models in Appendix.

\paragraph{Task-oriented Semantic Parsing}
The fourth evaluation task is cross-lingual task-oriented semantic parsing.
In this task, the input is a user utterance and the goal is to predict the intent of the utterance and fill the corresponding slots.
We conduct experiments on two recently proposed benchmarks: (i) mTOP \cite{li2020mtop} and (ii) mATIS++ \cite{xu-etal-2020-end}.
We jointly train the BERT models as suggested in \citet{chen2019bert}.

We summarize the evaluation task benchmark datasets and evaluation metrics in Table \ref{table:statitics}.

\subsection{Implementation Details}
We collect the universal part-of-speech tags and the dependency parse of sentences by pre-processing the datasets using UDPipe.\footnote{https://ufal.mff.cuni.cz/udpipe/2} 
We fine-tune mBERT on the pre-processed datasets and consider it as the baseline for our proposed syntax-augmented mBERT.
We extend the XTREME framework \cite{hu2020xtreme} that is developed based on \texttt{transformers} API \cite{wolf-etal-2020-transformers}.
We use the same hyper-parameter setting for mBERT models, as suggested in XTREME.
For the graph attention network (GAT), we set $L_G = 4, k = 4,$ and $d_g = 64$ (resulting in $\sim$0.5 million parameters).
We tune $\delta$\footnote{
We observed that the value of $\delta$ depends on the downstream task and the source language. For example, a larger $\delta$ value is beneficial for tasks taking a pair of text sequences as inputs, while a smaller $\delta$ value results in better performances for tasks taking single text input.
Experiments on PAWS-X using each target language as the source language indicate that $\delta$ should be set to a larger value for source language with longer text sequences (e.g., Arabic) and vice versa.
}
(shown in Eq. \eqref{eq:mask}) and $\alpha$ (weight of the tree structure prediction loss) in the range $[1, 2, 4, 8]$ and $[0.5 - 1.0]$, respectively.
We detail the hyper-parameters in the Appendix.

\begin{table*}[t]
\begin{subtable}{.5\textwidth}
  \centering
    \begin{tabular}{c|c c c c c c c}
    \hline
    $s_1$/$s_2$ & en & de & es & fr & ja & ko & zh \\
    \hline
    en	&	- 	&	0.7	&	1.6	&	1.4	&	\cellcolor{lgray}4.7	&	2.5	&	\cellcolor{lgray}5.4	\\
    de	&	0.5	    &	-	    &	2.0	&	2.1	    &	\cellcolor{lgray}5.1	&	\cellcolor{lgray}3.5	&	\cellcolor{lgray}5.9	\\
    es	&	1.0	&	2.1	&	-	&	1.7	    &	\cellcolor{lgray}4.6	&	3.0	&	\cellcolor{lgray}6.6	\\
    fr	&	0.9	&	1.7	&	1.9	    &   -	    &	\cellcolor{lgray}5.0	&	2.7	&	\cellcolor{lgray}5.4	\\
    ja	&	\cellcolor{lgray}5.2	&	\cellcolor{lgray}5.3	&	\cellcolor{lgray}5.6	&	\cellcolor{lgray}5.1	&	-	&	\cellcolor{lgray}5.9	&	\cellcolor{lgray}5.1	\\
    ko	&	3.1	&	2.8	&	\cellcolor{lgray}4.3	&	\cellcolor{lgray}3.9	&	\cellcolor{lgray}6.4	&	-	&	\cellcolor{lgray}5.1	\\
    zh	&	\cellcolor{lgray}5.8	&	\cellcolor{lgray}5.5	&	\cellcolor{lgray}6.3	&	\cellcolor{lgray}6.0	&	\cellcolor{lgray}6.1	&	\cellcolor{lgray}4.5	&   -		\\
    \hline
    \end{tabular}
  \caption{PAWS-X}
  \label{table:pawsx_xling}
\end{subtable}%
\begin{subtable}{.5\textwidth}
  \centering
    \begin{tabular}{c| c c c c c c c}
    \hline
    $q$/$c$	&	en	&	es	&	de	&	ar	&	hi	&	vi	&	zh	\\ \hline
    en	&	-	&	-0.2	&	0.3	&	0.4	&	0.9	&	0.6	&	1.1	\\ 
    es	&	\cellcolor{lgray}4.1	&	-	&	\cellcolor{lgray}3.5	&	\cellcolor{lgray}5.4	&	\cellcolor{lgray}5.3	&	\cellcolor{lgray}7.3	&	\cellcolor{lgray}7.6	\\
    de	&	\cellcolor{lgray}3.5	&	2.8	&	-	&	\cellcolor{lgray}4.0	&	2.9	&	\cellcolor{lgray}4.0	&	\cellcolor{lgray}5.0	\\
    de	&	1.8	&	2.4	&	1.1	&	-	&	-0.1	&	\cellcolor{lgray}6.2	&	\cellcolor{lgray}4.4	\\
    hi	&	1.0	&	1.8	&	0.5	&	0.2	&	-	&	-0.6	&	1.0	\\
    vi	&	\cellcolor{lgray}5.6	&	\cellcolor{lgray}4.5	&	\cellcolor{lgray}5.5	&	\cellcolor{lgray}6.9	&	\cellcolor{lgray}4.2	&	-	&	\cellcolor{lgray}5.5	\\
    zh	&	\cellcolor{lgray}3.8	&	\cellcolor{lgray}3.3	&	\cellcolor{lgray}4.4	&	2.4	&	0.9	&	\cellcolor{lgray}5.4	&	-	\\
    \hline
    \end{tabular}
  \caption{MLQA}
  \label{table:mlqa_xling}
\end{subtable} 
\caption{
The performance difference between syntax-augmented mBERT and mBERT in the \emph{generalized} cross-lingual transfer setting.
The rows and columns indicate (a) language of the first and second sentences in the candidate pairs and (b) context and question languages.
The gray cells have a value greater than or equal to the average performance difference, which is 3.9 and 3.1 for (a) and (b).
}
\label{table:xling}
\end{table*}

\section{Experiment Results}
We aim to address the following questions. 
\begin{compactenum}
\item Does augmenting mBERT with syntax improve (generalized) cross-lingual transfer?
\item Does incorporating syntax benefit specific languages or language families?
\item Which NLP tasks or types of tasks get more benefits from utilizing syntax?
\end{compactenum}

\subsection{Cross-lingual Transfer}
Experiment results to compare mBERT and syntax-augmented mBERT are presented in Table \ref{table:all_result}.
Overall, the incorporation of language syntax in mBERT improves cross-lingual transfer for the downstream tasks, in many languages by a significant margin  ($p<0.05$, t-test).
The average performances across all languages on XNLI, PAWS-X, MLQA, and mTOP benchmarks improve significantly (by at least 1 point). 
On the other benchmarks: Wikiann, CoNLL, XQuAD, and mATIS++, the average performance improvements are 0.5, 0.2, 0.8, and 0.7 points, respectively.
Note that the performance gains in the source language (English) for all the datasets except Wikiann is $\leq$ 0.3.
This indicates that cross-lingual transfer gains are not due to improving the downstream tasks, but instead, language syntax helps to transfer across languages.

\subsection{Generalized Cross-lingual Transfer}
In the generalized cross-lingual transfer setting \cite{lewis-etal-2020-mlqa}, the input text sequences for the downstream tasks (e.g., text classification, QA) may come from different languages.
As shown in Figure \ref{table:qa_ex1}, given the context passage in English, a multilingual QA model should answer the question written in Spanish.
Due to the parallel nature of the existing benchmark datasets: XNLI, PAWS-X, MLQA, and XQuAD, we evaluate mBERT and its' syntax-augmented variant on the generalized cross-lingual transfer setting.
The results for PAWS-X and MLQA are presented in Table \ref{table:xling} (results for the other datasets are provided in Appendix).

In both text classification and QA benchmarks, we observe significant improvements for most language pairs. 
In the PAWS-X text classification task, language pairs with different typologies (e.g., en-ja, en-zh) have the most gains. 
When Chinese (zh) or Japanese (ja) is in the language pairs, the performance is boosted by at least 4.5\%.
The dataset characteristics explain this; the task requires modeling structure, context, and word order information.
On the other hand, in the XNLI task, the performance gain pattern is scattered, and this is perhaps syntax plays a less significant role in the XNLI task.
The largest improvements result when the languages of the premise and hypothesis sentences belong to \{Bulgarian, Chinese\} and \{French, Arabic\}.

In both QA datasets, syntax-augmented mBERT boosts performance when the question and context languages are typologically different except the Hindi language.
Surprisingly, we observe a large performance gain when questions in Spanish and German are answered based on the English context.
Based on our manual analysis on MLQA, we suspect that although questions in Spanish and German are translated from English questions (by human), the context passages are from Wikipedia that often are not exact translation of the corresponding English passage. 
Take the context passages in Figure \ref{table:qa_ex1} as an example.
We anticipate that syntactic clues help a QA model in identifying the correct answer span when there are more than one semantically equivalent and plausible answer choices.

\begin{figure*}
\centering
\includegraphics[width=1.0\linewidth]{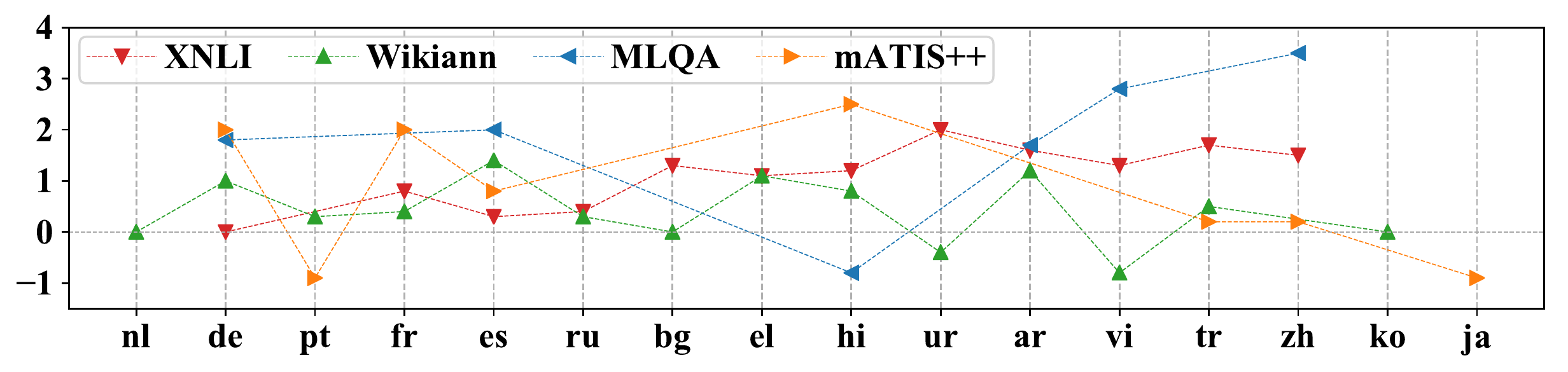}
\caption{
Performance improvements for XNLI, Wikiann, MLQA, and mATIS++ across languages.The languages in x-axis are grouped by language families: IE.Germanic (nl, de), IE.Romance (pt, fr, es), IE.Slavic (ru, bg), IE.Greek (el), IE.Indic (hi, ur), Afro-asiatic (ar, vi), Altaic (tr), Sino-tibetan (zh), Korean (ko), and Japanese (ja).
}
\label{figure:mtop}
\end{figure*}

\subsection{Analysis \& Discussion}
We discuss and analyze our findings on the following points based on the empirical results.

\paragraph{Impact on Languages}
We study if fine-tuning syntax-augmented mBERT on English (source language) impacts specific target languages or families of languages.
We show the performance gains on the target languages grouped by their families in four downstream tasks in Figure \ref{figure:mtop}.
There is no observable trend in the overall performance improvements across tasks.
However, the XNLI curve weakly indicates that when target languages are typologically different from the source language, there is an increase in the transfer performance (comparing left half to the right half of the curve).


\paragraph{Impact of Pre-training GAT}
Before fine-tuning syntax-augmented mBERT, we pre-train GAT on the 17 target languages (discussed in \cref{sec:fine-tuning}).
In our experiments, we observe such pre-training boosts semantic parsing performance, while there is a little gain on the classification and QA tasks.
We also observe that pre-training GAT diminishes the gain of fine-tuning with the auxiliary objective (predicting the tree structure).
We hypothesize that pre-training or fine-tuning GAT using auxiliary objective helps when there is limited training data.
For example, semantic parsing benchmarks have a small number of training examples, while XNLI has many. As a result, the improvement due to pre-training or fine-tuning GAT in the semantic parsing tasks is significant, and in the XNLI task, it is marginal.

\paragraph{Discussion}
To foster research in this direction, we discuss additional experiment findings.

\smallskip
\noindent$\bullet$ A natural question is, instead of using GAT, why we do not modify attention heads in mBERT to embed the dependency structure (as shown in Eq. \ref{eq:mask}).
We observed a consistent performance drop across all the tasks if we intervene in self-attention (blocking pair-wise attention).
We anticipate fusing GAT encoded syntax representations helps as it adds bias to the self-attention.
For future works, we suggest exploring ways of adding structure bias, e.g., scaling attention weights based on dependency structure \cite{bugliarello-okazaki-2020-enhancing}.

\smallskip
\noindent$\bullet$ Among the evaluation datasets, Wikiann consists of sentence fragments, and the semantic parsing benchmarks consist of user utterances that are typically short in length.
Sorting and analyzing the performance improvements based on sequence lengths suggests that the utilization of dependency structure has limited scope for shorter text sequences. 
However, part-of-speech tags help to identify span boundaries improving the slot filling tasks.   

\subsection{Limitations and Challenges}
In this work, we assume we have access to an off-the-shelf universal parser, e.g., UDPipe \cite{straka-strakova-2017-tokenizing} or Stanza \cite{qi-etal-2020-stanza} to collect part-of-speech tags and the dependency structure of the input sequences.
Relying on such a parser has a limitation that it may not support all the languages available in benchmark datasets, e.g., we do not consider Thai and Swahili languages in the benchmark datasets.

There are a couple of challenges in utilizing the universal parsers.
First, universal parsers tokenize the input sequence into words and provide part-of-speech tags and dependencies for them.
The tokenized words may not be a part of the input.\footnote{For example, in the German sentence ``Wir gehen zum kino'' (we are going to the cinema), the token ``zum'' is decomposed into words ``zu'' and ``dem''.}
As a result, tasks requiring extracting text spans (e.g., QA) need additional mapping from input tokens to words.
Second, the parser's output word sequence is tokenized into wordpieces that often results in inconsistent wordpieces resulting in degenerated performance in the downstream tasks.\footnote{This happen for languages, such as Arabic as parsers normalize the input that lead to inconsistent characters between input text and the output tokenized text.} 


\section{Related Work}

\paragraph{Encoding Syntax for Language Transfer}
Universal language syntax, e.g., part-of-speech (POS) tags, dependency parse structure, and relations are shown to be helpful for cross-lingual transfer 
\cite{kozhevnikov-titov-2013-cross, prazak-konopik-2017-cross, wu2017improved, subburathinam-etal-2019-cross, liu-etal-2019-neural-cross, zhang-etal-2019-syntax-enhanced-self, xie-etal-2020-contextual, ahmad2020gate}.
Many of these prior works utilized graph neural networks (GNN) to encode the dependency graph structure of the input sequences.
In this work, we utilize graph attention networks (GAT) \cite{velickovic2018graph}, a variant of GNN that employs the multi-head attention mechanism.

\paragraph{Syntax-aware Multi-head Attention}
A large body of prior works investigated the advantages of incorporating language syntax to enhance the self-attention mechanism \cite{vaswani2017attention}.
Existing techniques can be broadly divided into two types. 
The first type of approach relies on an external parser (or human annotation) to get a sentence's dependency structure during inference.
This type of approaches embed the dependency structure into contextual representations \cite{wu2017improved, chen-etal-2017-improved, wang2019source, wang-etal-2019-self-attention, zhang-etal-2019-syntax-enhanced-self, zhang2019sgnet, bugliarello-okazaki-2020-enhancing, sachan2020syntax, ahmad2020gate}.
Our proposed method falls under this category; however, unlike prior works, our study investigates if fusing the universal dependency structure into the self-attention of existing multilingual encoders help cross-lingual transfer.
Graph attention networks (GATs) that use multi-head attention has also been adopted for NLP tasks
\cite{huang-carley-2019-syntax} also fall into this category.
The second category of approaches does not require the syntax structure of the input text during inference.
These approaches are trained to predict the dependency parse via supervised learning \cite{strubell-etal-2018-linguistically, deguchi-etal-2019-dependency}.

\section{Conclusion}
In this work, we propose incorporating universal language syntax into multilingual BERT (mBERT) by infusing structured representations into its multi-head attention mechanism.
We employ a modified graph attention network to encode the syntax structure of the input sequences.
The results endorse the effectiveness of our proposed approach in the cross-lingual transfer.
We discuss limitations and challenges to drive future works.


\section*{Acknowledgments} 
We thank Yuqing Tang for his insightful comments on our paper and anonymous reviewers for their helpful feedback.
We also thank UCLA-NLP group for helpful discussions and comments.

\section*{Broader Impact}
In today's world, the number of speakers for some languages is in billions, while it is only a few thousands for many languages.
As a result, a few languages offer large-scale annotated resources, while for many languages, there are limited or no labeled data.
Due to this disparity, natural language processing (NLP) is extremely challenging in the low-resourced languages.
In recent years, cross-lingual transfer learning has achieved significant improvements, enabling us to avail NLP applications to a wide range of languages that people use across the world.
However, one of the challenges in cross-lingual transfer is to learn the linguistic similarity and differences between languages and their correlation with the target NLP applications.
Modern transferable models are pre-trained on unlabeled humongous corpora such that they can learn language syntax and semantic and encode them into universal representations.
Such pre-trained models can benefit from explicit incorporation of universal language syntax during fine-tuning for different downstream applications.
This work presents a thorough study to analyze the pros and cons of utilizing Universal Dependencies (UD) framework that consists of grammar annotations across many human languages.
Our work can broadly impact the development of cross-lingual transfer solutions and making them accessible to people across the globe.
In this work, we discuss the limitations and challenges in utilizing universal parsers to benefit the pre-trained models.
Among the negative aspects of our work is the lack of explanation that why some languages get more benefits over others due to universal syntax knowledge incorporation.

\bibliography{anthology,acl2021}
\bibliographystyle{acl_natbib}

\clearpage
\appendix

\twocolumn[{%
 \centering
 \Large\bf Supplementary Material: Appendices \\ [20pt]
}]

\section{Model Implementations}
We follow the standard way to model text classification, named entity recognition, and task-oriented semantic parsing using mBERT. 
However, since our proposed model uses the input sentences' dependency structure, we frame question answering (QA) as multi-sentence reading comprehension.
The input context is split into a list of sentences and train the mBERT model to predict the answer sentence and extract the answer span from the selected sentence following \citet{clark-gardner-2018-simple}.
We concatenate the question and each sentence from the context passage and use the [CLS] token representation to score the candidate sentences.
We adopt the \emph{shared-normalization} approach from the ``confidence method'' as suggested in \citet{clark-gardner-2018-simple} and pick the highest-scored sentence to extract the answer span during inference.
Our approach of utilizing syntax can be extended to apply to passages directly. 
To combine all the sentences' dependency structure in the passage, we can add edges from the [CLS] token to the roots of all the sentences' dependency tree.
However, would that approach work in practice requires empirical study, and we leave this as future work.

\section{Hyper-prameter Details}
We present the hyper-parameter details in Table \ref{tab:hparam_details}.
\begin{table*}[!htb]
\centering
\begin{tabular}{l|p{0.8\linewidth}}
\hline
\multicolumn{2}{l}{Graph Attention Network (GAT)} \\ 
\hline
\# layers ($L_G$) & 4 (tuned on [2, 4, 8]) \\
\# heads ($k$)  & 4 (tuned on [1, 2, 4, 8]) \\
$d_g$ & 64 (tuned on [32, 64, 128]) \\
$\delta$  & 4 (classification, QA), 1 (NER, semantic parsing) \\
\hline
\multicolumn{2}{l}{Syntax-augmented mBERT} \\ 
\hline
\# syntax-layers & 12 (tuned in the range 1 -- 12) \\
\# syntax-heads  & 1 (XNLI, PAWS-X, Wikiann, CoNLL, mATIS++), 2 (MLQA, XQuAD, mTOP) \\
$\alpha$ & 0.0 (XNLI), 0.2 (mATIS++), 0.5 (PAWS-X, Wikiann, MLQA, XQuAD), 1.0 (CoNLL), 2.0 (mTOP) \\
\# epochs & 3 (QA), 5 (classification), 10 (NER, semantic parsing) \\
\hline
\end{tabular}
\caption{
Details of the hyper-parameters used during fine-tuning syntax-augmented mBERT.
}
\label{tab:hparam_details}
\end{table*}

\section{Additional Experiment Results}

\paragraph{Cross-lingual Transfer}
We provide the exact match (EM) and F1 accuracy of the MLQA dataset in Table \ref{table:mlqa}.
Intent classification accuracy, slot F1, and exact match (EM) accuracy for the task-oriented semantic parsing is reported in Table zero-shot cross-lingual transfer results for the evaluation tasks in Table \ref{table:parsing_result}.
We highlight the cross-lingual transfer gap for mBERT and syntax-augmented mBERT on the evaluation tasks in Table \ref{table:xl_gap}.


\begin{table*}[!ht]
\centering
\resizebox{\linewidth}{!}{%
\begin{tabular}{l@{\hskip 0.05in}|c@{\hskip 0.1in} c@{\hskip 0.1in} c@{\hskip 0.1in} c@{\hskip 0.1in} c@{\hskip 0.1in} c@{\hskip 0.1in} c@{\hskip 0.05in}|c}
\hline
models & en & es & de & ar & hi & vi & zh & avg \\
\hline 
\citeauthor{lewis-etal-2020-mlqa} & 77.7/65.2 & 64.3/46.6 & 57.9/44.3 & 45.7/29.8 & 43.8/29.7 & 57.1/38.6 & 57.5/37.3 & 57.7/41.6 \\
mBERT (ours) & 80.5/67.2 & 63.9/44.1 & 59.0/43.3 & 47.2/28.6 & 47.5/32.1 & 56.5/35.2 & 57.8/33.0 & 58.9/40.5 \\ 
mBERT+Syn. & 80.4/67.3 & 65.9/47.1 & 60.8/45.1 & 48.9/30.3 & 46.7/32.4 & 59.3/38.0 & 60.1/35.5 & 60.3/42.2 \\
\hline
\end{tabular}
}
\caption{
Zero-shot cross-lingual transfer performance (F1/EM) of mBERT on MLQA dataset. 
}
\label{table:mlqa}
\end{table*}

\begin{table*}[t]
\centering
\begin{tabular}{l|c c c| c c c| c c c| c c c}
\hline
& \multicolumn{6}{c}{mTOP \cite{li2020mtop}} & \multicolumn{6}{c}{mATIS++ \cite{xu-etal-2020-end}} \\
\cline{2-13}
& \multicolumn{3}{c|}{mBERT (ours)} & \multicolumn{3}{c|}{mBERT+Syn.} & \multicolumn{3}{c|}{mBERT (ours)} & \multicolumn{3}{c}{mBERT+Syn.} \\
\hline
en & 95.5 & 90.0 & 81.0 & 95.6 & 90.2 & 81.3 & 97.3 & 94.9 & 86.0 & 97.3 & 94.9 & 86.2 \\
fr & 63.8 & 63.4 & 38.8 & 67.8 & 64.1 & 41.2 & 92.9 & 70.2 & 36.9 & 90.4 & 74.1 & 38.9 \\
es & 68.7 & 62.1 & 40.2 & 73.1 & 62.9 & 43.0 & 94.1 & 74.0 & 43.7 & 90.8 & 77.2 & 44.5 \\
de & 58.2 & 60.2 & 28.1 & 63.2 & 59.4 & 30.0 & 89.7 & 68.2 & 38.1 & 89.5 & 69.4 & 40.1 \\
hi & 41.2 & 30.7 & 9.8 & 44.2 & 31.4 & 11.5 & 80.4 & 49.4 & 16.2 & 80.4 & 54.7 & 18.7 \\
ja & - & - & - & - & - & - & 83.6 & 60.7 & 28.2 & 81.9 & 61.3 & 27.3 \\
pt & - & - & - & - & - & - & 94.8 & 66.3 & 38.2 & 92.7 & 66.7 & 37.3 \\
tr & - & - & - & - & - & - & 71.3 & 16.9 & 1.3 & 68.7 & 18.6 & 1.5 \\
zh & - & - & - & - & - & - & 87.6 & 24.3 & 7.8 & 86.0 & 21.2 & 8.0 \\ \hline
Avg. & 65.5 & 61.3 & 39.6 & 68.8 & 61.6 & 41.4 & 88.0 & 58.3 & 32.9 & 86.4 & 59.8 & 33.6 \\
\hline
\end{tabular}
\caption{
Zero-shot cross-lingual task-oriented semantic parsing results. 
The values for each model indicates intent accuracy, slot F1, and exact match, respectively.
}
\label{table:parsing_result}
\end{table*}

\begin{table*}[!ht]
\centering
\begin{tabular}{l|c@{\hskip 0.1in} c|c@{\hskip 0.1in} c|c@{\hskip 0.1in} c|c@{\hskip 0.1in} c}
\hline
Model & XNLI & PAWS-X & Wikiann & CoNLL & MLQA & XQuAD & mTOP & mATIS++ \\
\hline
mBERT (ours) & 15.5 & 12.8 & 16.1 & 17.2 & 25.4 & 23.2 & 51.8 & 59.7 \\
mBERT+Syn. & 14.2 & 11.3 & 15.7 & 16.9 & 23.5 & 22.1 & 49.9 & 59.2 \\
\hline
\end{tabular}
\caption{
The cross-lingual transfer gap of mBERT and syntax-augmented mBERT on the evaluation tasks. The transfer gap is the difference between performance on the English test set and the other languages' average performance.
A transfer gap of 0 indicates perfect cross-lingual transfer. 
For the QA datasets, we use F1 scores.
}
\label{table:xl_gap}
\end{table*}

\paragraph{Generalized Cross-lingual Transfer}
In generalized cross-lingual transfer, we assume the task inputs are a pair of text that belong to two different languages, e.g., answering Spanish question based on an English context \cite{lewis-etal-2020-mlqa}.
We present the generalized cross-lingual transfer performance of syntax-augmented mBERT on XNLI, MLQA, and XQuAD in Table \ref{table:pawsx_xling}, \ref{table:xnli_xling_result}, \ref{table:mlqa_xling_appendix}, and \ref{table:xquad_xling}, respectively.
The performance differences between syntax-augmented mBERT and mBERT on the generalized cross-lingual transfer on XNLI and XQuAD is presented in Table \ref{table:xnli_xling_gap} and \ref{table:xquad_xling_diff}.

\begin{figure*}
\centering
\includegraphics[width=0.6\linewidth]{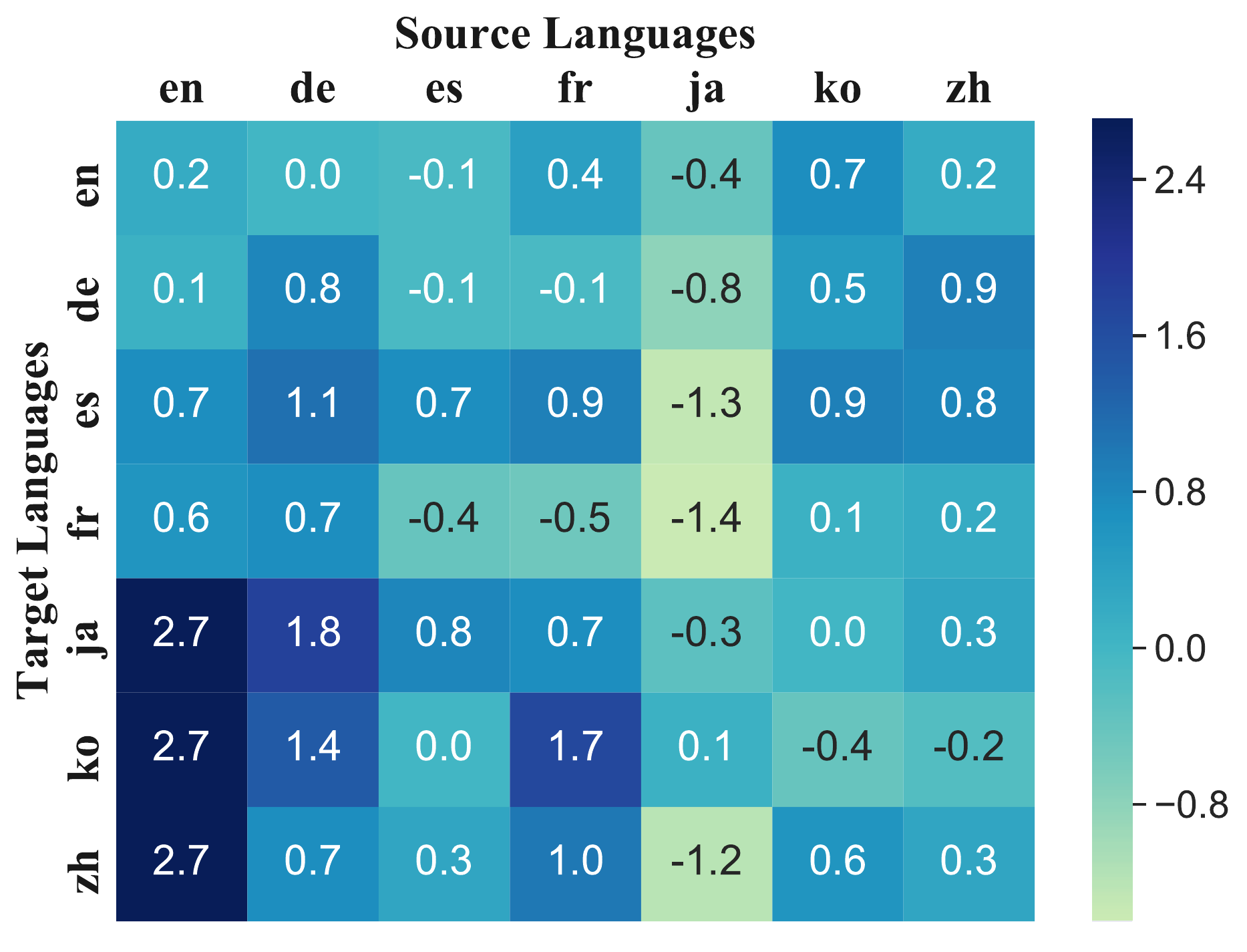}
\caption{
Zero-shot cross-lingual transfer performance difference between syntax-augmented mBERT and mBERT for PAWS-X task using different languages as source.
}
\label{figure:multi_pawsx}
\end{figure*}

\paragraph{Different Source Languages}
In our study, we primarily use English as the source language as training examples used in all the benchmarks are in English.
However, authors of many of these benchmarks released translated-train examples in the target languages.
This allows us to train mBERT and syntax-augmented mBERT in different languages (as source) and examine how it impacts cross-lingual transfer. 
We perform experiments on PAWS-X task and present the results in Figure \ref{figure:multi_pawsx}.
We observe the largest transfer performance improvements when English and German are used as the source language.
The improvements are relatively smaller when Japanese, Korean, and Chinese languages are used as the source language.
We suspect that the dependency parser may not accurately parse translated sentences, and as a result, we do not see an explainable trend in the improvements.

\begin{table*}[!ht]
\centering
\begin{tabular}{l c c c c c c c}
\hline
$s_1$/$s_2$	&	en	&	de	&	es	&	fr	&	ja	&	ko	&	zh	\\ \hline
en	&	-	&	85.6	&	87.2	&	86.3	&	66.1	&	68	&	70.5	\\
de	&	86.8	&	-	&	82	&	82	&	65	&	67.7	&	68.6	\\
es	&	87.2	&	82	&	-	&	85.5	&	63.7	&	66.1	&	68.3	\\
fr	&	86	&	81.8	&	84.8	&	-	&	64.2	&	66.6	&	67.9	\\
ja	&	65.6	&	64.5	&	64.6	&	64.2	&	-	&	67.3	&	68	\\
ko	&	69.3	&	67.3	&	67.9	&	67.7	&	69	&	-	&	66.4	\\
zh	&	71.4	&	68.1	&	68.8	&	68.9	&	67.5	&	65.9	&	-	\\
\hline
\end{tabular}
\caption{
Generalized cross-lingual transfer performance of syntax-augmented mBERT on PAWS-X.
The row and column indicates the language of the input sentence pairs.
}
\label{table:pawsx_xling}
\end{table*}
\begin{table*}[!ht]
\centering
\begin{tabular}{l c c c c c c c c c c c c c}
\hline
p/h	&	en	&	fr	&	es	&	de	&	ru	&	el	&	bg	&	ar	&	tr	&	hi	&	ur	&	vi	&	zh	\\ \hline
en	&	-	&	70.1	&	70.1	&	66.2	&	64.8	&	57.3	&	61.8	&	59.1	&	53.8	&	53.5	&	51.2	&	64.4	&	65.2	\\
fr	&	72.5	&	-	&	69	&	63.9	&	63.2	&	56.5	&	60.2	&	58.6	&	52.5	&	52.1	&	49.9	&	62.4	&	61.3	\\
es	&	72.5	&	68.6	&	-	&	63	&	63.7	&	57.7	&	60.8	&	59.1	&	52.6	&	51.5	&	48.5	&	61.5	&	60.9	\\
de	&	71.1	&	65.7	&	65.1	&	-	&	63.1	&	56.1	&	60	&	58	&	52.8	&	53.2	&	50.6	&	60.3	&	60.5	\\
ru	&	69.3	&	64.5	&	65.5	&	62.5	&	-	&	55.9	&	62.7	&	57	&	51	&	51.2	&	48.2	&	58.8	&	58.5	\\
el	&	63	&	59.8	&	61	&	57.5	&	56.9	&	-	&	56.9	&	55.9	&	50.6	&	49.8	&	47.8	&	56.6	&	54.7	\\
bg	&	68.4	&	63	&	64.6	&	61.2	&	64	&	57	&	-	&	57.4	&	51	&	52.5	&	48.1	&	58.8	&	59.1	\\
ar	&	63.7	&	59.2	&	59.9	&	56.6	&	55.8	&	53.5	&	54.2	&	-	&	50.1	&	50.9	&	49.7	&	55.8	&	55.5	\\
tr	&	60	&	55.2	&	54.9	&	53.9	&	51.9	&	51.7	&	52.5	&	53.2	&	-	&	50.4	&	48.4	&	53.3	&	53.7	\\
hi	&	61.1	&	55	&	54.7	&	54.7	&	53.7	&	52	&	52.3	&	53.7	&	50	&	-	&	53.2	&	54	&	53.7	\\
ur	&	59.9	&	55.1	&	54.1	&	53.5	&	50.7	&	49.9	&	49	&	52.9	&	48.6	&	54.6	&	-	&	50.7	&	52.6	\\
vi	&	65.9	&	60.2	&	59.3	&	56.3	&	55.5	&	53.2	&	52.7	&	54.2	&	47.8	&	49.7	&	46.8	&	-	&	62.3	\\
zh	&	66.8	&	58.9	&	58.4	&	56.1	&	54.8	&	50.9	&	53.4	&	54.5	&	48.8	&	49.4	&	47.2	&	61.5	&	-	\\
\hline
\end{tabular}
\caption{
Generalized cross-lingual transfer performance of syntax-augmented mBERT on XNLI.
The row and column indicates the language of premise and hypothesis.
}
\label{table:xnli_xling_result}
\end{table*}

\begin{table*}[!ht]
\centering
\begin{tabular}{l|c c c c c c c}
\hline
q/c	&	en	&	es	&	de	&	ar	&	hi	&	vi	&	zh	\\ \hline
en	&	80.4	&	67.6	&	63.1	&	53.3	&	55.1	&	64.0	&	59.9	\\
es	&	69.3	&	65.9	&	58.1	&	47.8	&	46.2	&	56.0	&	52.2	\\
de	&	69.2	&	62.9	&	60.8	&	49.5	&	50.6	&	56.2	&	52.3	\\
ar	&	47.0	&	44.1	&	41.3	&	48.9	&	35.3	&	38.9	&	40.8	\\
hi	&	41.6	&	36.5	&	35.8	&	32.4	&	46.7	&	35.1	&	33.8	\\
vi	&	56.2	&	49.7	&	47.5	&	40.6	&	39.8	&	59.3	&	47.9	\\
zh	&	58.5	&	52.0	&	49.0	&	41.0	&	38.8	&	53.0	&	60.1	\\
\hline
\end{tabular}
\caption{
F1 score for generalized cross-lingual transfer of syntax-augmented on MLQA.  
Columns show context language, rows show question language.
}
\label{table:mlqa_xling_appendix}
\end{table*}

\begin{table*}[!ht]
\centering
\begin{tabular}{l|c c c c c c c c c c}
\hline
q/c	&	en	&	es	&	de	&	ru	&	el	&	ar	&	hi	&	tr	&	vi	&	zh	\\ \hline
en	&	84.2	&	74.4	&	70.7	&	65.1	&	59.4	&	52.6	&	52.9	&	53.1	&	63.9	&	51.7	\\
es	&	67.1	&	71.3	&	58.0	&	55.6	&	49.0	&	45.1	&	41.9	&	41.9	&	48.9	&	39.5	\\
de	&	66.6	&	59.9	&	69.2	&	56.0	&	50.7	&	43.3	&	45.6	&	41.9	&	50.6	&	41.7	\\
ru	&	63.7	&	61.6	&	58.0	&	69.9	&	49.1	&	42.8	&	45.9	&	41.5	&	51.0	&	40.4	\\
el	&	48.6	&	46.7	&	42.9	&	41.7	&	63.4	&	33.9	&	37.2	&	30.4	&	36.9	&	28.1	\\
ar	&	47.0	&	47.4	&	41.3	&	42.3	&	37.6	&	55.6	&	37.7	&	29.1	&	33.9	&	31.7	\\
hi	&	39.1	&	36.7	&	37.1	&	33.9	&	28.8	&	29.2	&	56.2	&	29.7	&	31.2	&	24.2	\\
tr	&	38.7	&	35.2	&	33.9	&	31.1	&	26.9	&	25.1	&	25.9	&	49.2	&	26.2	&	21.6	\\
vi	&	53.5	&	49.0	&	44.2	&	43.9	&	38.2	&	34.3	&	36.5	&	33.7	&	64.2	&	37.5	\\
zh	&	54.1	&	48.8	&	45.9	&	45.6	&	36.6	&	37.6	&	38.3	&	35.5	&	48.7	&	58.0	\\
\hline
\end{tabular}
\caption{
F1 score for generalized cross-lingual transfer for XQuAD.  
Columns show context language, rows show question language.
}
\label{table:xquad_xling}
\end{table*}

\begin{table*}[!ht]
\centering
\begin{tabular}{l c c c c c c c c c c c c c}
\hline
p/h & en & fr & es & de & ru & el & bg & ar & tr & hi & ur & vi & zh \\ \hline
en & - & 0.9 & 0.2 & 0.6 & 0.5 & 0.6 & 0.6 & 0.9 & 1.2 & 0.8 & 1.1 & 1.2 & 1.3 \\
fr & -0.5 & - & 1.1 & \cellcolor{lgray}1.6 & 1.3 & 1.2 & \cellcolor{lgray}1.6 & \cellcolor{lgray}2.3 & \cellcolor{lgray}1.8 & \cellcolor{lgray}1.6 & 0.9 & \cellcolor{lgray}2.0 & 1.4 \\
es & -0.4 & \cellcolor{lgray}1.5 & - & \cellcolor{lgray}1.5 & 0.5 & 0.9 & \cellcolor{lgray}1.5 & \cellcolor{lgray}1.9 & \cellcolor{lgray}1.9 & \cellcolor{lgray}1.7 & 1.0 & \cellcolor{lgray}1.7 & \cellcolor{lgray}1.6 \\
de & 0.3 & \cellcolor{lgray}1.7 & 1.2 & - & 0.7 & 0.7 & 1.4 & \cellcolor{lgray}2.2 & \cellcolor{lgray}1.7 & 1.0 & 0.8 & 1.2 & 1.2 \\
ru & 0.8 & \cellcolor{lgray}2.2 & 1.4 & 1.4 & - & \cellcolor{lgray}1.5 & 1.2 & \cellcolor{lgray}2.2 & \cellcolor{lgray}2.0 & \cellcolor{lgray}1.6 & \cellcolor{lgray}1.7 & \cellcolor{lgray}1.8 & \cellcolor{lgray}1.8 \\
el & 0.7 & \cellcolor{lgray}2.2 & \cellcolor{lgray}1.8 & 1.1 & 1.1 & - & \cellcolor{lgray}1.8 & \cellcolor{lgray}1.8 & \cellcolor{lgray}1.7 & 1.4 & 1.4 & 1.4 & 1.3 \\
bg & 1.1 & \cellcolor{lgray}3.5 & \cellcolor{lgray}2.6 & \cellcolor{lgray}2.3 & 0.1 & 1.4 & - & \cellcolor{lgray}2.3 & \cellcolor{lgray}1.7 & \cellcolor{lgray}2.1 & \cellcolor{lgray}1.5 & \cellcolor{lgray}2.9 & \cellcolor{lgray}2.6 \\
ar & 0.8 & \cellcolor{lgray}2.7 & \cellcolor{lgray}1.8 & \cellcolor{lgray}2.1 & 1.3 & 1.2 & \cellcolor{lgray}1.8 & - & \cellcolor{lgray}2.0 & \cellcolor{lgray}1.5 & \cellcolor{lgray}1.6 & \cellcolor{lgray}1.8 & \cellcolor{lgray}1.6 \\
tr & 0.6 & \cellcolor{lgray}2.4 & \cellcolor{lgray}2.0 & \cellcolor{lgray}1.9 & 1.1 & \cellcolor{lgray}1.5 & \cellcolor{lgray}1.5 & \cellcolor{lgray}1.6 & - & \cellcolor{lgray}1.7 & \cellcolor{lgray}1.5 & \cellcolor{lgray}1.7 & 1.4 \\
hi & 0.9 & \cellcolor{lgray}1.5 & \cellcolor{lgray}1.5 & 1.4 & 0.9 & 0.6 & 1.4 & \cellcolor{lgray}1.5 & 1.3 & - & 1.0 & 1.1 & 1.2 \\
ur & 1.0 & \cellcolor{lgray}1.9 & \cellcolor{lgray}1.7 & 1.4 & 0.4 & 0.6 & 0.6 & 1.2 & 1.3 & \cellcolor{lgray}1.6 & - & 0.7 & 1.0 \\
vi & 0.8 & \cellcolor{lgray}3.1 & \cellcolor{lgray}2.0 & \cellcolor{lgray}2.0 & 1.4 & 1.4 & \cellcolor{lgray}1.8 & \cellcolor{lgray}2.4 & 1.4 & 1.2 & 1.3 & - & \cellcolor{lgray}2.1 \\
zh & \cellcolor{lgray}1.7 & \cellcolor{lgray}2.8 & \cellcolor{lgray}2.3 & \cellcolor{lgray}2.2 & \cellcolor{lgray}1.5 & 1.3 & \cellcolor{lgray}2.2 & \cellcolor{lgray}1.9 & \cellcolor{lgray}1.6 & 1.4 & \cellcolor{lgray}1.6 & \cellcolor{lgray}3.0 & - \\
\hline
\end{tabular}
\caption{
Cross-lingual transfer performance difference between syntax-augmented mBERT and mBERT on the XNLI dataset in the \emph{generalized} setting.
The row and column indicates the language of premise and hypothesis. The gray cell have a value $\geq$ 1.5  (average difference).
}
\label{table:xnli_xling_gap}
\end{table*}
\begin{table*}[!ht]
\centering
\begin{tabular}{l|c c c c c c c c c c}
\hline
q/c	&	en	&	es	&	de	&	ru	&	el	&	ar	&	hi	&	tr	&	vi	&	zh	\\ \hline
en	&	-0.3	&	-1.0	&	-0.3	&	0.9	&	2.2	&	2.4	&	1.6	&	1.3	&	1.7	&	0.0	\\
es	&	\cellcolor{lgray}4.4	&	1.6	&	\cellcolor{lgray}3.7	&	\cellcolor{lgray}4.3	&	\cellcolor{lgray}4.4	&	\cellcolor{lgray}5.6	&	\cellcolor{lgray}6.1	&	\cellcolor{lgray}4.3	&	\cellcolor{lgray}7.4	&	\cellcolor{lgray}4.5	\\
de	&	\cellcolor{lgray}4.4	&	\cellcolor{lgray}3.9	&	2.3	&	\cellcolor{lgray}3.1	&	2.7	&	\cellcolor{lgray}4.3	&	2.4	&	\cellcolor{lgray}5.4	&	\cellcolor{lgray}4.5	&	\cellcolor{lgray}3.4	\\
ru	&	1.8	&	0.2	&	1.6	&	-0.3	&	1.2	&	\cellcolor{lgray}4.7	&	-0.6	&	\cellcolor{lgray}3.5	&	2.2	&	\cellcolor{lgray}4.1	\\
el	&	-0.2	&	0.8	&	1.3	&	\cellcolor{lgray}4.2	&	0.2	&	\cellcolor{lgray}4.5	&	0.4	&	\cellcolor{lgray}3.1	&	\cellcolor{lgray}4.1	&	3.0	\\
ar	&	0.6	&	-0.1	&	2.4	&	1.4	&	1.1	&	0.8	&	-0.3	&	\cellcolor{lgray}5.2	&	\cellcolor{lgray}3.6	&	0.9	\\
hi	&	1.1	&	1.6	&	-1.5	&	2.0	&	\cellcolor{lgray}4.2	&	2.2	&	-1.2	&	0.2	&	2.2	&	2.7	\\
tr	&	\cellcolor{lgray}3.3	&	\cellcolor{lgray}4.3	&	\cellcolor{lgray}4.0	&	\cellcolor{lgray}3.8	&	\cellcolor{lgray}5.3	&	\cellcolor{lgray}5.9	&	2.4	&	0.6	&	\cellcolor{lgray}4.1	&	\cellcolor{lgray}4.7	\\
vi	&	\cellcolor{lgray}4.5	&	\cellcolor{lgray}5.1	&	\cellcolor{lgray}6.9	&	\cellcolor{lgray}4.8	&	\cellcolor{lgray}5.7	&	\cellcolor{lgray}6.5	&	\cellcolor{lgray}3.3	&	\cellcolor{lgray}3.7	&	\cellcolor{lgray}3.5	&	\cellcolor{lgray}5.3	\\
zh	&	\cellcolor{lgray}4.0	&	\cellcolor{lgray}4.4	&	\cellcolor{lgray}5.4	&	\cellcolor{lgray}3.7	&	\cellcolor{lgray}5.3	&	\cellcolor{lgray}5	&	1.6	&	\cellcolor{lgray}3.8	&	\cellcolor{lgray}3.6	&	-1.2	\\
\hline
\end{tabular}
\caption{
F1 score difference for generalized crosslingual transfer for XQuAD. Columns show context language, rows show question language. The gray cells have a value $\geq$ 3.1 (average difference).
}
\label{table:xquad_xling_diff}
\end{table*}

\end{document}